\documentclass[journal]{IEEEtran}
\usepackage{graphicx}
\usepackage[cmex10]{amsmath}
\usepackage{array}
\usepackage{url}
\usepackage{amssymb,theorem,float,bbm,bm}
\usepackage{enumerate}
\usepackage{setspace}

\def\1{\mathbbm{1}}

{\theorembodyfont{\rmfamily} }
{\theorembodyfont{\rmfamily} }

\begin{document}

\title{Virus Detection in Multiplexed Nanowire Arrays using Hidden Semi-Markov models}

\author{Shalini~Ghosh,
  Patrick Lincoln,
  Christian Petersen,
  Alfonso Valdes\\
  Computer Sciences Laboratory, SRI International, CA.
}
\maketitle

\begin{abstract}
In this paper, we address the problem of real-time detection of
viruses docking to nanowires, especially when multiple viruses dock to
the same nano-wire. The task becomes more complicated when there is an
array of nanowires coated with different antibodies, where different
viruses can dock to each coated nanowire at different binding
strengths. We model the array response to a viral agent as a pattern
of conductance change over nanowires with known modifier --- this
representation permits analysis of the output of such an array via
belief network (Bayes) methods, as well as novel generative models
like the Hidden Semi-Markov Model (HSMM).
\end{abstract}

\section{Introduction}
\label{sec:intro}

Real-time detection of viruses in the field of healthcare and
biodefense has become a very important problem in recent times. In
this work, we follow the methodology of real-time electrical detection
of viruses with nanowire field-effect transistors described in
Patolsky et al.~\cite{patolsky04}. In this method, nanowires are coated
with antibodies of a particular kind of virus. The main idea is as
follows: if that type of virus is present in the environment, then the
virus molecules would dock with the antibody molecules on the nanowire
and change the conductance of the nanowire. Signals of the nanowire
conductance as a function of time are typically analyzed to figure out
whether the virus has docked to the nanowire, thereby detecting the
presence of the virus.

If only a single virus docks to the nanowire, then the problem is
relatively simple: one just has to figure out whether or not there is
a binding event of the virus to the nanowire. However, in practice,
apart from the main virus which docks to the nanowire, multiple
viruses from the same family can also exhibit weak binding to the
coated nanowire. The problem then becomes more challenging, since now
the task is not only to detect a binding event but also to identify
which virus molecule has docked. The task becomes further complicated
when there is an array of nanowires coated with different antibodies
where different viruses can dock to each coated nanowire at different
binding strengths.

Building on Patolsky et al.~\cite{patolsky04}, we propose belief
network and generative probabilistic models for detecting virus
docking in large nanowire arrays.  Patolsky et al. note in their paper
that there are limitations in their method and other methods in the
detection of rapidly mutating, engineered, and/or new viruses. One
approach to overcome this limitation, they note, is through
multiplexed nanowire arrays, including nanowires modified with general
viral-cell surface detectors and antibody libraries (collectively,
modifiers).  Herein we propose concepts to enable this enhanced
detection capability.  We seek to develop approaches to address the
following questions:

\begin{itemize}

\item Given the output of a multiplexed array with nanowires modified
with a known set of modifiers, what is the content of the analyzed
sample?  Variants of this include: Is a given virus present, and what
is the most likely mix of viruses? Stressing factors could include
failures of similarly modified nanowires as well as contaminants in
the sample.

\item Given a desired list of viral agents of interest, what is an
effective approach to selecting modifiers so as to detect and
distinguish the agents of interest?  Factors to consider include
achieving adequate redundancy to tolerate nanowire failures as well as
a selection of modifiers that covers the space of viral agents of
interest.

\end{itemize}

Our approaches are based on the representation of the array response
to a viral agent as a pattern of conductance change over nanowires
with known modifiers. This representation permits analysis of the
output of such an array via belief network (Bayes) methods, as well as
generative models like the Hidden Semi-Markov Model
(HSMM)~\cite{murphy02}. Our development explicitly comprehends
``noisy'' output from the array, caused by faulty nanowires or
contaminants.

This paper is organized as follows.  We first discuss the expected
response of single nanowires to viral particles, as presented in the
reference, with schematics of idealized conductance traces as well as
traces of actual data.  We then present approaches to detecting an
expected signal on a single nanowire, building from a simple threshold
approach to matched filtering~\cite{blackledge06} on conditioned data,
and finally presenting a histogram detrending technique to address
detection in some of the more challenging traces from Patolsky et
al.~\cite{patolsky04}. We then describe virus detection in a notional
multiplexed nanowire array.  In such an array the nanowires are
divided into sets with all nanowires within a set modified with the
same modifier (providing redundant detection capability of response to
a specific modifier), and different sets of nanowires are modified
with antibodies from an antibody library (providing coverage of a
breadth of viral agents of interest). Subsequently, we propose a more
advanced model of virus detection, the Hidden Semi-Markov Model, and
show experimental results with the HSMM.  Finally, we give an
assessment of the cross-correlation in the nanowire array. Note that
in this paper, we will occasionally refer to the method in Patolsky et
al.~\cite{patolsky04} as the Patolsky method.

\section{Nanowire Response to Viral Binding}
\label{sec:response}

\begin{figure}[hbtp]
\begin{center}
\includegraphics[width=9cm]{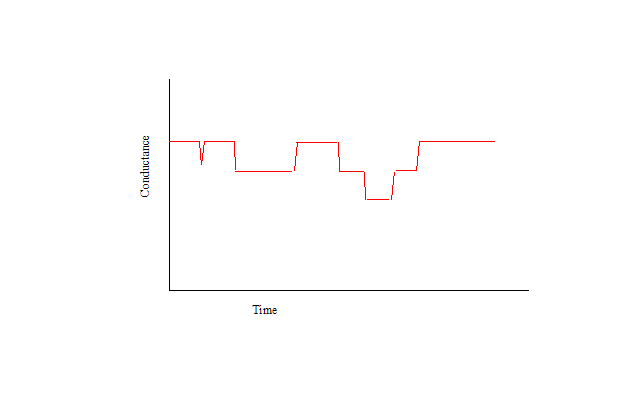}
\end{center}
\caption{Idealized conductance trace showing, respectively, nonspecific
transient spike, specific binding, and superimposed specific bindings.}
\label{fig1}
\end{figure}

\begin{figure}[hbtp]
\begin{center}
\includegraphics[width=9cm]{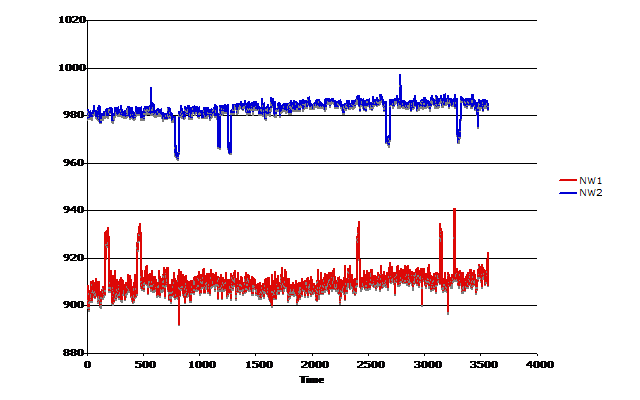}
\end{center}
\caption{Traces for Influenza and Adenovirus (corresponds to data
from Figure 4 of Patolsky et al.~\cite{patolsky04}).}
\label{fig2}
\end{figure}

A nanowire modified with specific antibody receptors responds to viral
bindings as follows.  Viral particles sampling but not binding to a
receptor site on the wire cause a transient change in conductance,
which manifests as a spike of brief duration in a trace of conductance
versus time.  Transient spikes are also observed coincident with
fluidic injections, and these spikes may correlate across nanowires in
a multiplexed array so as to allow removal of noise features
correlated across wires.  Specific bindings appear as a ``boxcar''
change in conductance, either positive or negative, depending on
solution pH and the charge of the modifier.  For a given
concentration, the duration of the boxcar in time is fairly
consistent, as is the amplitude change.  The observed response
corresponds to single viral particle bindings, and the interval of
conductance change corresponds to a particle attaching, remaining
bound for some time interval, and then releasing, as confirmed by
electron micrograph.  Some cases of multiple bindings to the same wire
have been observed by Patolsky et al.; these appear as superimposed
boxcars in the trace.  Figure~\ref{fig1} represents an idealized
conductance trace for a single nanowire, showing a transient spike
feature corresponding to a nonspecific contact, a boxcar corresponding
to a specific binding, and a superimposed boxcar corresponding to a
specific binding followed by a second specific binding before the
first particle releases. Figure~\ref{fig2} shows an actual trace of
data from the Patolsky method (this figure corresponds to
Figure~\ref{fig4} in Patolsky et al.~\cite{patolsky04}).  There are
two nanowires, the first modified with anti-Influenza Type A antibody
(NW1) and the second with anti-adenovirus group III (NW2). The figure
shows expected specific bindings as the respective viral agents are
introduced. The noise characteristics and boxcar shapes are typical of
what we have observed, although these traces do not exhibit
significant trends.

\section{Matched Filter Approach to Single Wire Detection}

Matched filtering describes a variety of techniques whereby a
representation of a signal of interest is convolved with a series or
image in which the signal may or may not be present.  Regions that
agree well with the matched filter will correspond to local maxima in
the matched filter output.  These local maxima are declared to be
instances of the signal of interest if they are above some
threshold. For reasons of computational efficiency, both signal and
data are transformed, typically using a Fast Fourier Transform
(FFT)~\cite{blackledge06}. Convolution corresponds to pointwise
multiplication in transform space.  Matched filtering is an optimal
detection algorithm in the case of additive signal in white noise.
Much of the efficacy in the approach depends on conditioning of the
data so as to satisfy this assumption.  Common noise sources include:

\begin{itemize}

\item Uncorrelated noise, which is often effectively removed by
subtracting the data mean and dividing by the standard deviation.

\item Frequency content (in time or space), removed by estimation of
and then dividing out the power spectrum.

\item Trends, which must be estimated and removed.

\item Signal capture, in which the signal is sufficiently strong so as
to significantly influence estimates of noise, frequency content, or
trend.  For example, in typical conductance traces in our case, the
magnitude and duration of amplitude changes due to specific bindings
are sufficient to affect sample statistics.

\end{itemize}

\begin{figure}[hbtp]
\begin{center}
\includegraphics[width=9cm]{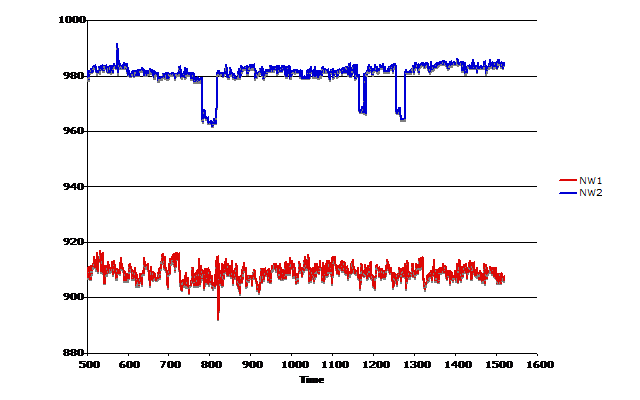}
\end{center}
\caption{Subtrace of 1024 points from Patolsky et al.}
\label{fig3}
\end{figure}

For simplicity, our initial matched filter analysis considered only
uncorrelated noise. As such, the data normalization procedure is
simply to subtract the grand mean and divide by the standard
deviation.  This simple procedure achieves a degree of data whitening.
A matched filter is an idealization of the hypothesized signal one is
trying to detect in the data.  In this case, the matched filter is a
manually constructed -20 nS boxcar of duration 20 seconds.  For
FFT~\cite{blackledge06} indexing reasons, this is ``unwrapped'' to
comprise 10 points of value -20 in the first 10 positions and 10
points of value -20 in the last 10 positions of an array 1024 long,
which is otherwise zero.  The matched filter is approximately whitened
by dividing by the standard deviation of the conductance data.  We
focus on the interval from 500 to 1523 seconds, which provides 1024
points and is a convenient array size for Fourier analysis. This
partial trace is shown in Figure~\ref{fig3}.

\begin{figure}[hbtp]
\begin{center}
\includegraphics[width=9cm]{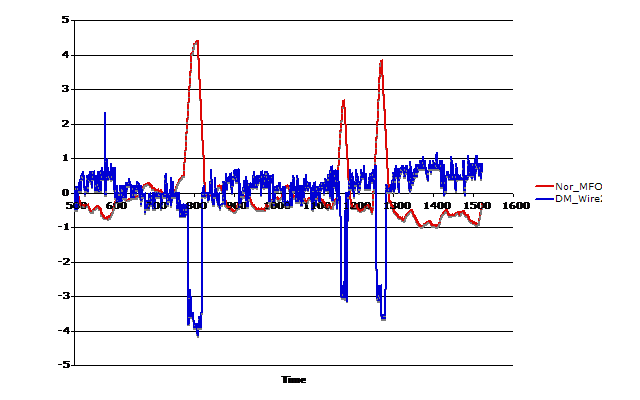}
\end{center}
\caption{Normalized conductance trace with matched filter output.}
\label{fig4}
\end{figure}

Over this interval, NW1 exhibits spike features only, so we focus on
NW2, which exhibits one spike and three boxcars.  The whitened data
and matched filter are processed through a Fast Fourier Transform
(FFT).  The convolution of the matched filter and the data is achieved
in transform space as element-wise multiplication of the Fourier
series.  An inverse transform produces the matched filter output,
which is then normalized by dividing by its own standard deviation (it
is in theory zero mean -- the mean we obtained here was 1.8e-11).
Figure~\ref{fig4} shows the normalized data trace for wire 2 as well
as the matched filter output.  Peaks in the matched filter output
correspond to regions of maximal match between the data and the
filter.  Although this is a somewhat simple case, we observe as
expected three matched filter peaks corresponding to the bindings of
interest, with no matched filter ringing at the transient spike or
anywhere else along the trace.

\section{Histogram Detrending and Advanced Detection}

\begin{figure}[hbtp]
\begin{center}
\includegraphics[width=9cm]{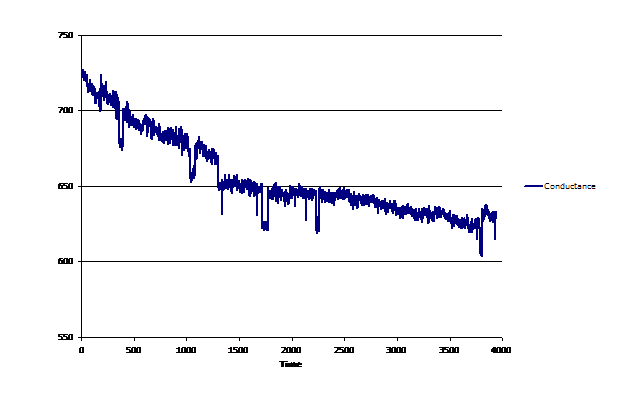}
\end{center}
\caption{Influenza + Parmoxyvirus.}
\label{fig5}
\end{figure}

\begin{figure}[hbtp]
\begin{center}
\includegraphics[width=9cm]{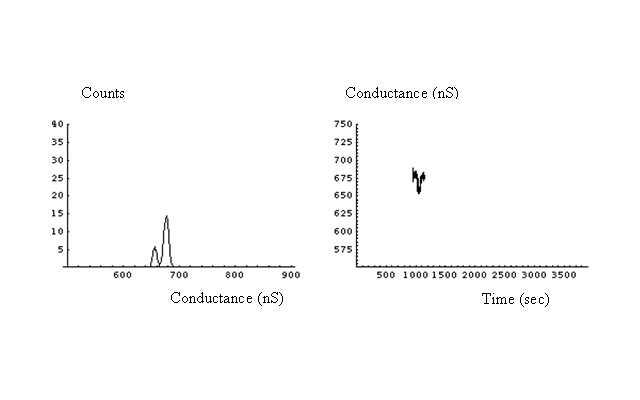}
\end{center}
\caption{Influenza + Parmyoxovirus histogram and data window containing virus
bind/unbind signature.}
\label{fig6}
\end{figure}

\begin{figure}[hbtp]
\begin{center}
\includegraphics[width=9cm]{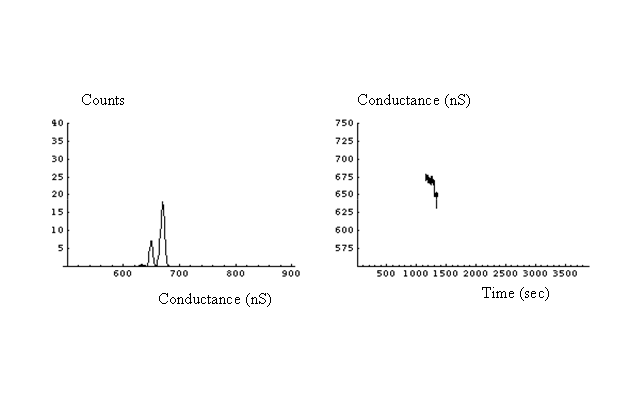}
\end{center}
\caption{Influenza + Parmyoxovirus histogram and data window containing
possible virus sequential bind signature -- local mean at a
conductance of roughly 675 nS.}
\label{fig7}
\end{figure}

\begin{figure}[hbtp]
\begin{center}
\includegraphics[width=9cm]{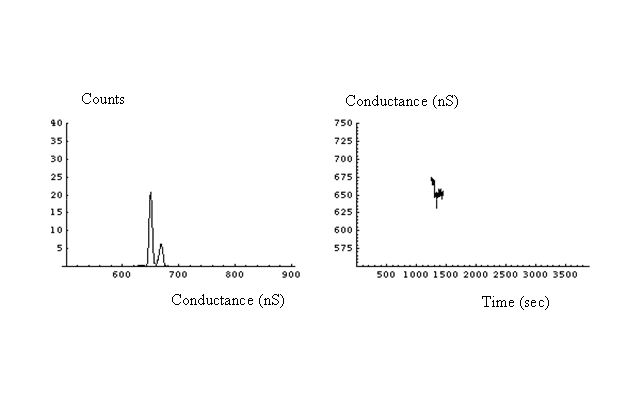}
\end{center}
\caption{Influenza + Parmyoxovirus histogram and data window containing
possible virus sequential bind signature -- local mean now at a
conductance of 645 nS.}
\label{fig8}
\end{figure}

Figure~\ref{fig5} represents unpublished data from Patolsky et al.,
showing a nanowire modified for Influenza detection, in an experiment
where first Influenza only and then a mixture of Influenza and
Parmyoxovirus are introduced.  The latter introduction is immediately
before the break in the trend (at time 1300).  The
influenza/parmyoxovirus data shows a significant nonlinear trend and
also contains signal transitions that deviate from the boxcar model
(possibly the result of sequential virus binding/unbinding).  Data
such as this requires more sophisticated processing than presented
above in order to estimate and remove the trend.  Three detrending
approaches were tried with this data.  The first method is a
conventional curve fitting method in which we fit a global function to
the entire data set and subtract that function from the data (assuming
that the fitted function represents the data trend).  The global
fitting approach turned out to be undesirable because the change in
conductance at approximately 1300 seconds produces a persistent shift
in the data mean and biases the trend locally.  Another approach is to
model the trend locally using a moving average. The trend and bias
removal is better than for the previous fitted trend removal case.
However, the moving average still shows some variation near binding
events.  Nonetheless, this result is much more appropriate for other
modeling (e.g., auto-regression) and detection techniques (matched
filtering).  Finally, a more sophisticated approach is taken in which
a histogram is utilized to obtain a more accurate representation of
the local mean.  In this approach, a histogram is computed in much the
same way as the moving average.  In this case, histograms are
successively computed over a 200 second moving window.  Each histogram
represents a density function of the conductance values within the
window.  Since the data is largely characterized by slowly moving
trends and boxcar or step functions, the histograms will typically
exhibit one or two distinct populations, the latter in the case that
the window contains significant points from the background and the
signal.  In Figure~\ref{fig6}, the histogram data window overlaps a
virus bind/unbind signature in the data.  Since the width of the
bind/unbind signature is roughly 10-15\% of the window width (20-30
second pulse width in a 200 second window), the histogram population
corresponding to the minimum in the signature is somewhat smaller than
the population corresponding to the local data mean, resulting in two
distinct histogram modes.  Thus the local mean can be identified with
little biasing from the signature.  In Figures~\ref{fig7} and
\ref{fig8}, an apparent sequential virus bind occurs, shifting the
local mean.  In this case the relative heights of the two populations
in the histogram become comparable and then the maximum changes from
one population to the other as the histogram window moves in time
(that is, the modes corresponding to the binding and the background
are reversed).

\begin{figure}[hbtp]
\begin{center}
\includegraphics[width=9cm]{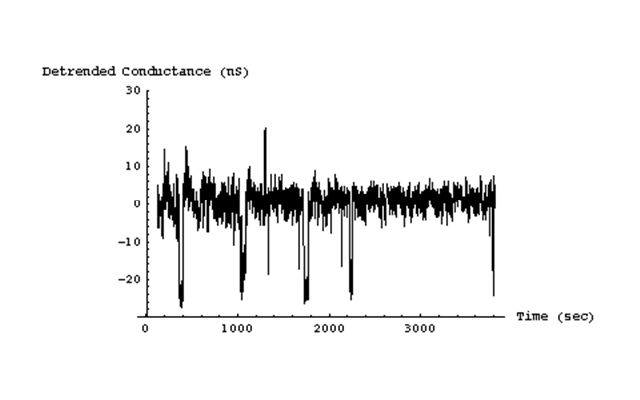}
\end{center}
\caption{Influenza + Parmyoxovirus conductance data with histogram
model trend removed.}
\label{fig9}
\end{figure}

By using the maximum value of the largest population in the histogram,
we can estimate the mean value of the local window in the data.  Step
changes in the data will be regarded as sudden changes in the mean as
opposed to a bind or unbind signature unless additional measures are
taken. We process the original data by choosing the value of
conductance for the maximum count value of each histogram as the local
mean at that point.  Then we move the window and repeat the
computation for each point in the original time series.  The original
time series data detrended by the model is shown in Figure~\ref{fig9}.

With the histogram model, the bias and trend removal is more stable
near binding events.  From a qualitative point of view, the signal to
noise ratio seems slightly greater with the histogram model.  One
disadvantage with this method is that the single step in the data at
1300 seconds appears as a spike rather than as a boxcar function --
additional analysis might be necessary to detect these transitions if
that were desirable.  Another disadvantage of this method is that it
is computationally much more expensive than the others.  However, this
method can be applied incrementally as data is gathered, so the cost
may not be significant.

\section{Detection in Notional Multiplexed Nanowire Arrays}

\begin{figure}[hbtp]
\begin{center}
\includegraphics[width=9cm]{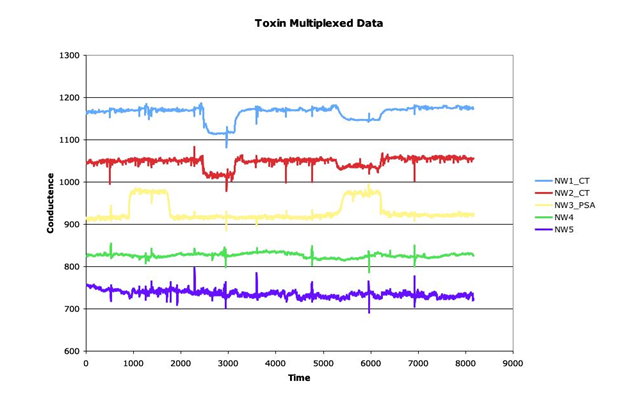}
\end{center}
\caption{5-wire multiplex array modified for CT and PSA.}
\label{fig10}
\end{figure}

Patolsky et al. has proposed multiplexed arrays for more reliable
detection of viral agents (reliability achieved through redundant
nanowires similarly modified) as well as detection of diverse agents
with a single array.  As reported in the paper, the team had built
arrays with fewer than 10 wires and limited redundancy, with the
eventual objective of scaling to arrays of hundreds or thousands of
nanowires.   Figure~\ref{fig10} shows traces obtained by Patolsky et
al. from a multiplexed array.  Here, wires 1 and 2 are modified with
the Cholera Toxin antibody (CT), while wire 3 is modified with the PSA
antibody.  Wires 4 and 5 are modified with ethanolamine only and serve
as controls.

\begin{figure}[hbtp]
\begin{center}
\includegraphics[width=9cm]{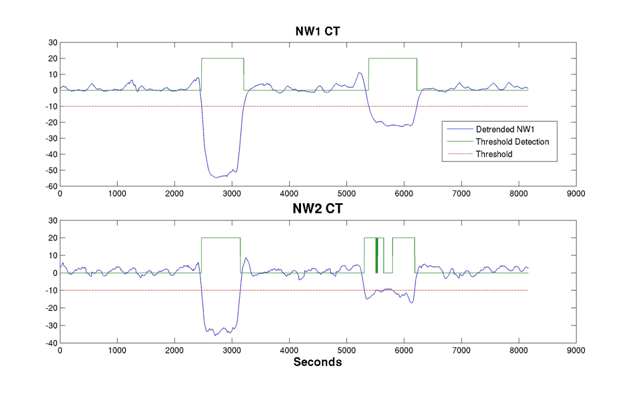}
\end{center}
\caption{Threshold detection in multiplex array.}
\label{fig11}
\end{figure}

Multiplexed arrays will consist of multiple nanowires modified with
the same antibody, drawn from an antibody library.  Ideally, an
antibody is specific to a viral agent.  In practice, we expect to see
some weak bindings (lower amplitude conductance change and/or shorter
duration of binding) in the case of an antibody specific to a virus
variant from the same family as the viral agent being analyzed in a
given run.  For example, an adenovirus variant is expected to exhibit
strong bindings to wires modified with antibodies specific to the
variant, and weaker binding to some wires modified with antibodies
specific to other adenovirus variants.  Bindings to wires modified for
other viral families (for example, adenovirus bindings to wires
modified with influenza antibodies) are expected to be rare.  A simple
detection approach that might prove robust for some types of nanowire
data is simple threshold testing.  In testing for substantial viral or
chemical concentrations (as opposed to single virus or particle
detection), one can expect binding-event signals such as in
Figure~\ref{fig10} lasting tens to hundreds of seconds as compared to
noise events ($\approx$ 1 second).  In such a case, low-pass filtering can
remove a large percentage of electrical noise and spiky ``sampling''
events on the nanowires.  The resulting signal can be tested against a
threshold to detect the presence of a binding event.  As an example,
we use the CT antibody nanowires NW1 and NW2 from the above data set.
Applying a low-pass filter to the NW1 and NW2 data results in the
signatures shown in Figure~\ref{fig11}.  Applying a threshold on the
negative-going waveform produces detections of the binding events.
The result from NW2 demonstrates that care must be taken in setting
the threshold.  One trades detection sensitivity against the
likelihood of false alarms in noisy data.  If the threshold is set too
close to the origin, the algorithm will interpret excursions in the
noise as detections.  A threshold too far away from the origin will
miss lower amplitude conductance changes due to lower concentration
samples.

The appropriate filtering and threshold value will depend on the
requirements for the particular application.  A separate threshold
would most likely be necessary for each wire in an array - awkward for
very large arrays.  However, the threshold could be adaptively set
depending on the background noise.  A one-time calibration phase where
the instrument is run on a neutral sample (to determine noise levels)
or subjected to actual samples would allow calibration of the
algorithm.  It would also be possible to dynamically adjust thresholds
(and thus false alarm rate or sensitivity) while the instrument runs
by maintaining some instantaneous measure of background noise for each
nanowire.  At the very least, individual thresholds would be necessary
to accommodate junction variations from the fabrication process.  An
advantage of the threshold technique is that it is simple and very
well understood.  It is also very easy to pass event width and height
information along for more processing -- for instance, determining the
concentration of a virus as well as its presence or combining the
signature of several similarly treated nanowires.  A disadvantage to
the threshold approach is susceptibility to noise.  The matched filter
or correlation filter techniques are much more resistant to
non-Gaussian noise because they would depend on the overall ``boxcar''
shape of the signature.  However, the width and height of binding
signatures can change depending on solution concentration and pH.  The
matched filter approach would require a bank of matched filters for
various ``boxcar'' dimensions or some normalization calculation to map
the signatures to a standard width or height.

\section{Integrating the Response on a Multiplex Array}

A future multiplexed detector will likely consist of an array of
nanowires, treated to respond to different agents or families of
agents, with the array containing similarly treated wires (replicates)
for robustness. We may represent the response of an array to a
particular virus as a pattern over the multiplexed array.  The
expected response patterns can be represented in tabular form as
idealized patterns.

Table 1 represents idealized response patterns for different agents
injected into a hypothetical array.  Each row of the table represents
one or a set of identically modified nanowires. The columns give the
notional response of the array (1.0 being the maximal strength
response) to respective viral agents.  This response could be, for
example, the normalized output of the matched filter or histogram
detector described previously, integrated over multiple wires in the
array.

We consider two families, the first with three variants, the second
with two.  The shorthand ``A1'' denotes the first variant of virus
family A, ``A2'' the second variant of the same family, and so on.  The
shorthand ``Anti-A1-1'' denotes the first member of the antibody library
specific for A1.  Note that for some viruses (for example, A3) we
have only one specific antibody in the library.  Expected (specific)
bindings have table entries close to unity; bindings within the
correct family have intermediate values, and bindings outside the
family have values near zero.

The last row in the table represents modification of nanowires with a
viral cell surface receptor.  All viruses are expected to have a
strong binding to this modifier.  In addition to columns corresponding
to specific viral agents, we include a column for a virus from the ``A''
family for which we have no specific antibody (Other-A) as well as a
new virus from an unknown family.  The former would have a weak
response to the ``A'' family antibodies, near zero response to the ``B''
family antibodies, and a strong response to the cell surface detector.
The latter would have a near zero response to all but the cell surface
detector.  The buffer would have a near zero response to all wires in
the array.

This representation enables various computational approaches to infer
the presence of one or more viruses given a noisy nanowire array
output (which will be in the form of a column vector of responses to
the various modifiers).  We describe a Bayes approach, in which we
treat the above responses as probabilities, even if they are not so in
the strict sense, and outline an applicable Bayes formalism.

\section{Naive Bayes Formalism}

Expressing the response on a 0 to 1 scale allows us to manipulate
responses as probabilities.  Specifically, we may computationally
treat the response in a particular table cell as the conditional
probability of a binding to the row variable given the presence of the
viral agent corresponding to the column.  As a specific example, we
consider that the conditional probability of a response for Anti-A1-1
given that A1 is present is 0.98.  This is written as
$P(\mbox{Anti-A1-1}=Yes|A1=present) = 0.98$.

%Table 1. Notional multiplex array response 

The Bayes formalism makes inference about hypotheses at a ``parent''
node based on evidence observed at ``leaf'' nodes or inferred at
intermediate nodes.  The inference engine uses the conditional
probability table (CPT), which expresses the algebraic relationship of
observable evidence to underlying hypotheses.  There is a CPT
corresponding to each arc in the belief network.  
%The CPT on the arc
%relating the response for Anti-A1-1 to the hypotheses corresponding to
%viral agents present is given in Table 2. 
It is assumed that the
hypothesis space is exhaustive and exclusive; that is, all possible
viral agents are enumerated, and there is no overlap.  This is part of
the motivation for including the ``New'' and ``Buffer'' hypotheses.

A Bayes net segment, in this case a subtree, consists of a parent node
enumerating the hypotheses {A1, A2, A3, Other-A, B1, B2, New,
Buffer} and leaf nodes for each antibody, with a CPT 
% analogous to Table 2 
on each arc.  An alternate structure encodes the hypotheses
{Virus, Buffer} at the root.  An intermediate node encodes hypotheses
{Family-A, Family-B, New, and Buffer}.

% Table 2. CPT relating response Anti-A1-1 to candidate hypotheses

We simulated the response of an array to mixtures consisting of a
known viral agent, a mix of known viral agents, an unknown viral agent
from a known family, and an unknown viral agent.

\section{Simulation Results: Single Viral Agent}

\begin{figure}[hbtp]
\begin{center}
\includegraphics[width=9cm]{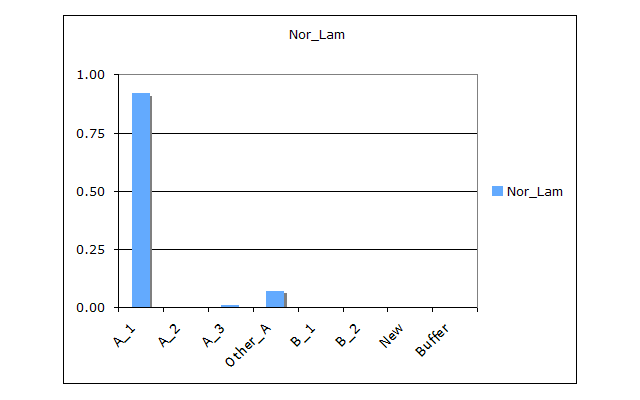}
\end{center}
\caption{Bayes response to viral agent A1.}
\label{fig12}
\end{figure}

We simulated the introduction of viral agent A1. Figure~\ref{fig12}
gives the output of the Bayes approach (in this case, a Bayes
posterior probability with the assumption of uniform prior
probabilities, for simplicity).  As expected, the response is
strongest for A1, with a much weaker response for a nonspecific agent
from the ``A'' family.

\section{Simulation Results: Mix of Viral Agents}

\begin{figure}[hbtp]
\begin{center}
\includegraphics[width=9cm]{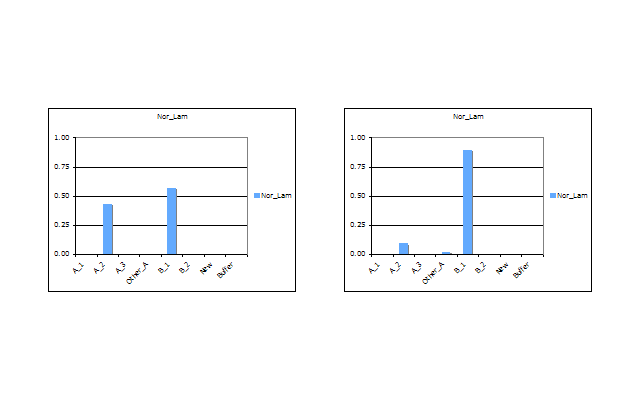}
\end{center}
\caption{Response to A2/B1 mix.}
\label{fig13}
\end{figure}

We simulated the introduction of A2 and B1.  The simulated output from
the array exhibited variability due to noise in the response.  Two
representative outputs are shown in the Figure~\ref{fig13}.  On the
left, we observe an output in which A2 and B1 dominate all other
outputs and are comparable to each other.  This would be the nominal
response in the absence of simulation noise.  On the right, we have an
output where B1 greatly dominates, with the response to A2 much lower
and the response to a nonspecific agent from the ``A'' family lower
still.

\section{Simulation Results: Unknown Viral Agent from Family A}

\begin{figure}[hbtp]
\begin{center}
\includegraphics[width=9cm]{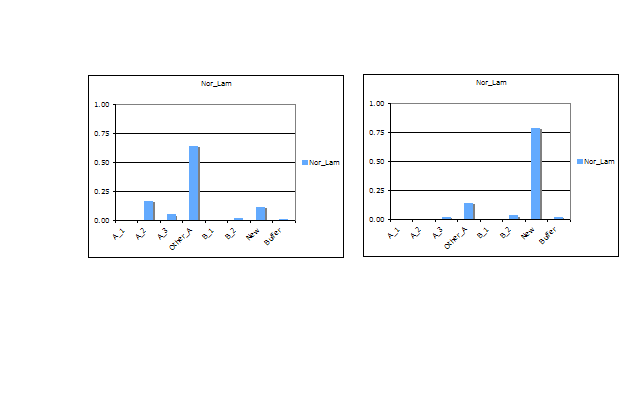}
\end{center}
\caption{Array response to unknown viral agent from family "A".}
\label{fig14}
\end{figure}

We next simulated the introduction of an unknown viral agent from
family ``A''.  There is no specific antibody for this agent, but we
expect a weak response to other antibodies from the same family.  This
response will be significantly weaker than the near-unity response
expected from a specific binding, but higher than the near-zero
response expected to antibodies for agents outside the family.
Moreover, we expect a strong response to the cell surface detector.
The result on the left in Figure~\ref{fig14} agrees with intuition,
with the agent ``Other-A'' most likely, but some responses for other
agents in the ``A'' family as well as a response for ``New''.
However, as with the mixture case, other realizations yield different
results.  Just one alternative is shown in the right panel, where the
dominant hypothesis is a new viral agent not from family ``A''.

\section{Simulation Results: New Viral Agent}

\begin{figure}[hbtp]
\begin{center}
\includegraphics[width=9cm]{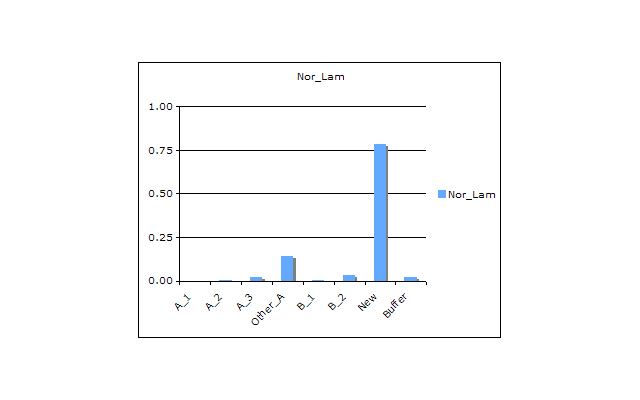}
\end{center}
\caption{Array response to new viral agent from unknown family.}
\label{fig15}
\end{figure}

The next result simulates the introduction of a new viral agent, from
neither family ``A'' nor ``B''.  The expectation is a strong binding
to the cell surface detector.  Figure~\ref{fig15} reflects the
expected result. This particular result is stable for different
simulation trials.

\section{Summary of Bayes Approach}

The preceding results mostly agree with intuition, but with some
exceptions.  The noise models underlying the situation are ad hoc, but
we observe ambiguity in the case of competing bindings (the mixture of
known viral agents in Figure~\ref{fig13}) or nonspecific bindings
(Figure~\ref{fig14}).

It is essential to characterize the noise in the response from
multiplexed arrays.  This noise will be reduced as similarly modified
nanowires are replicated in greater numbers.  However, we should not
assume the noise reduction that would be obtained from independent
replicates because the wires may respond similarly for some underlying
common mode effect (e.g., a process artifact for a batch of wires
similarly modified).

We also require more experimental data to obtain better estimates of
the strength of nonspecific bindings within a particular virus family.
It may be that these are weaker than what we have assumed here on our
zero-to-one scale.  In that case, it may prove difficult to
distinguish a new viral agent from a family for which we have some
antibodies from a viral agent from a new family altogether.

\section{Hidden Semi-Markov Model}

\begin{figure}[hbtp]
\begin{center}
\includegraphics[width=9cm]{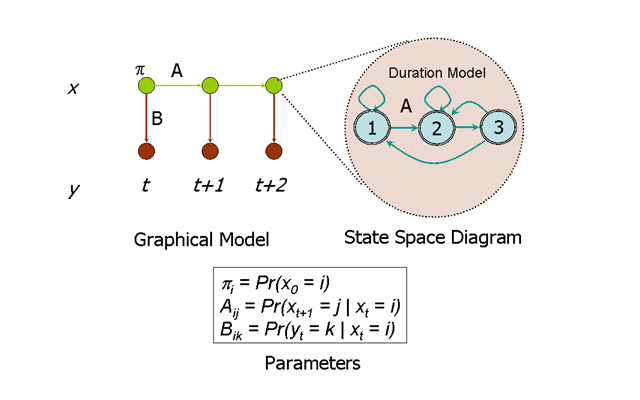}
\end{center}
\caption{Graphical model, state space and parameterization of a Hidden
Semi-Markov Model.}
\label{fig16}
\end{figure}

In an alternative approach, we create an underlying generative model
of the virus detection process, modeling both the strength and
duration of docking in the normalized detrended data. The
probabilistic model that we have considered is the hidden semi-Markov
model (HSMM)~\cite{duong05,murphy02}.  In the HSMM that we have used
there are two latent states: one state corresponds to the virus
docking to the nanowire, whereas the other state corresponds to the
virus not docking to the wire. There is a probability of going from
each hidden state to the other, and also a probability distribution
for the length of staying at a particular hidden state.
Figure~\ref{fig16} shows the graphical model and state-transition
diagram of an HSMM.

The main difference between an HSMM~\cite{murphy02} and a standard
hidden Markov model (HMM)~\cite{rabiner89} is in the length
distribution of staying at a particular hidden state. In an HMM this
distribution is geometric and hence falls off exponentially with
time. This may not be an appropriate model for virus docking, because
typically viruses would dock to a nanowire for a particular length of
time depending on the underlying biological process, and the observed
binding time would be distributed probabilistically (due to noise
factors) around this mean length of time. In an HSMM, an arbitrary
probability distribution can be used for modeling the length for which
the model stays at a particular hidden state. So, using an HSMM, we
can have a more realistic probabilistic model corresponding to the
time period for which a virus docks to a nanowire.

\begin{figure}[hbtp]
\begin{center}
\includegraphics[width=9cm]{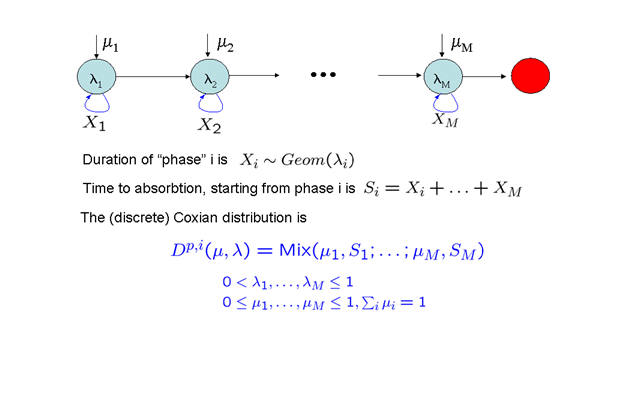}
\end{center}
\caption{Markov model representation of multiphase Coxian
distribution~\cite{duong05}.}
\label{fig17}
\end{figure}

In our HSMM, the output of each state follows a multinomial
distribution while the probability of staying in a particular state
follows a Coxian distribution~\cite{duong05}. The multinomial
distribution is generalization of the binomial distribution and
follows the multinomial theorem -- the multinomial distribution is
used in the observation model since we discretize the nanowire
conductance data as an initial simple noise-filtering step. The Coxian
distribution is a powerful mixture model-based duration model. It is a
mixture of the sum of exponential distributions -- we use this
distribution for the length model since Coxian is a multimodal
distribution, which makes it general enough to model dockings of
different lengths by different viruses to a
nanowire. Figure~\ref{fig17} shows the generative model of an M-phase
Coxian distribution, where the duration of phase k can be expressed as
the time to absorption in a Markov chain of M states when starting
from state k. The duration modeling is an important component of
generative modeling of virus docking -- staying in the docking state
for a short duration represents a noisy spike, while staying for too
long may represent an abnormality (e.g., a broken wire); staying for a
particular time duration in the docking process typically represents a
proper docking event.

\section{Simulation Experiments}

Following previous methodology, we first do median detrending of the
data followed by data whitening. This is followed by the thresholded
box-matched filtering algorithm explained earlier. The thresholded
output from the filtering algorithm is used to generate the training
labels for the HSMM. Using the labeled training data (which can be
potentially noisy if the thresholding is not perfect), the HSMM
parameters are learned. Finally, the learned HSMM is used to predict
whether or not there has been a virus detection in a given unlabeled
signal.

\begin{figure}[hbtp]
\begin{center}
\includegraphics[width=9cm]{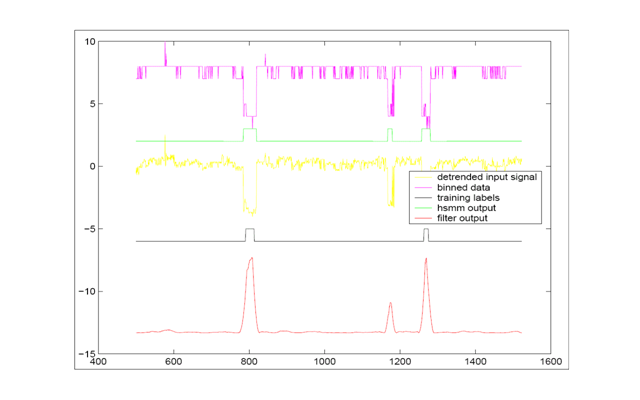}
\end{center}
\caption{Results on virus multiplexing data with only significant events as
training data for HSMM.}
\label{fig18}
\end{figure}

\begin{figure}[hbtp]
\begin{center}
\includegraphics[width=9cm]{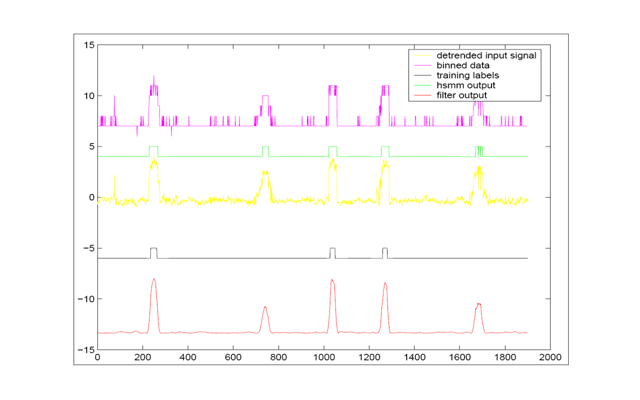}
\end{center}
\caption{Results on adenovirus data with only significant events as training
data for HSMM.}
\label{fig19}
\end{figure}

\begin{figure}[hbtp]
\begin{center}
\includegraphics[width=9cm]{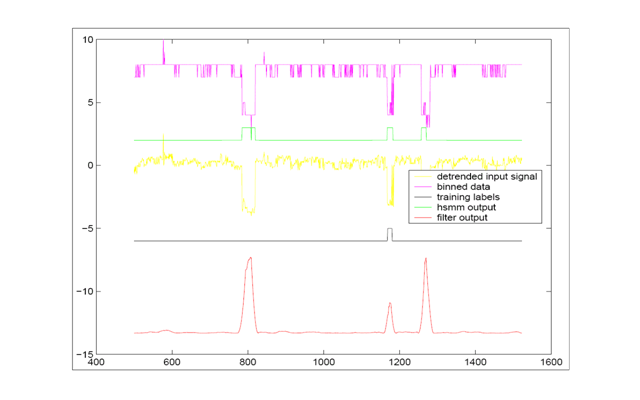}
\end{center}
\caption{Results on virus multiplexing data with only weak docking events as
training data for HSMM.}
\label{fig20}
\end{figure}

\begin{figure}[hbtp]
\begin{center}
\includegraphics[width=9cm]{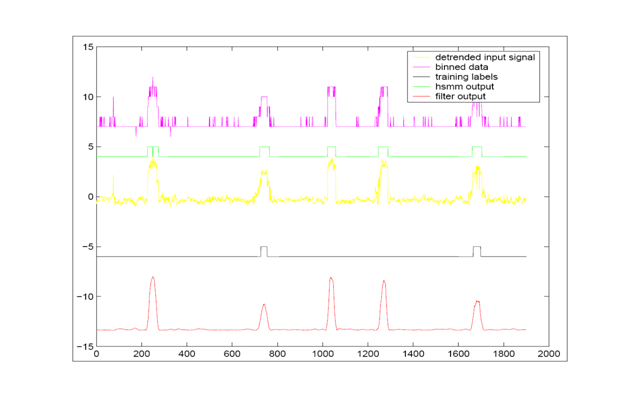}
\end{center}
\caption{Results on adenovirus data with only weak docking events as training
data for HSMM.}
\label{fig21}
\end{figure}

We have performed two experiments by injecting two kinds of noise in
the data. In the first experiment, noise is injected in the training
labels so that only the prominent docking events are labeled. The
results of these experiments are plotted in Figures~\ref{fig18} and
\ref{fig19}. In the second experiment, only the weak docking events
are used to generate docking labels. The results of the experiments
are plotted in Figures~\ref{fig20} and \ref{fig21}. In both these
experiments, the learned model could recover the correct labels of all
the docking events even though the training data was provided for a
subset of the labels, demonstrating that the HSMM can learn
effectively from limited and potentially noisy data. In the figures
for each dataset in each experiment, there are multiple plots -- from
top to bottom they are:

\begin{enumerate}

\item The discretized data, binned after detrending and whitening the
raw input data, given as input to the HSMM.

\item The output labels predicted by the HSMM (high corresponds to
docking, low implies nondocking). As can be seen from the plots, the
learned model is able to identify all the dockings correctly.

\item The raw data after median detrending and whitening.

\item The training labels given as input to the HSMM. This labeled
training data is used to learn the parameters of the HSMM. As
explained earlier, we intentionally gave incomplete labels to test the
robustness of the HSMM. As can be seen from the HSMM output in
Figure~\ref{fig19}, the HSMM was able to recover the docking regions
for which no training labels were given.

\item Output of boxcar filtering. We implemented FFT-based boxcar
filtering with a 20/20 width. The training labels in
Figure~\ref{fig21} were created by thresholding this signal.

\end{enumerate}

The HSMM method has the following advantages:

\begin{itemize}

\item The HSMM can learn with a small amount of labeled data. As we
see from the figures, we have provided incomplete labels as training
data but our model gives the correct prediction for all docking sites
for both datasets.

\item The generative model gives a natural underlying process model of
protein docking. We used a 2-phase Coxian distribution to model the
duration of the virus binding to the nanowires. If required, we can
use more number of phases to better model the underlying biological
process of the virus docking to the nanowire.

\end{itemize}

On these datasets, the generative HSMM has comparable performance to
the filtering and thresholding algorithm. On datasets where choosing a
single threshold after boxcar filtering is difficult (e.g., multiple
viruses docking to a nanowire), rich generative models like HSMM have
a clear benefit -- the HSMM can easily handle multiple dockings by
extending the HSMM output model to a mixture of multinomials, instead
of having a single multinomial.

\section{Assessment of Cross-Correlation of Noise Processes Across Nanowire Array}

\begin{figure}[hbtp]
\begin{center}
\includegraphics[width=9cm]{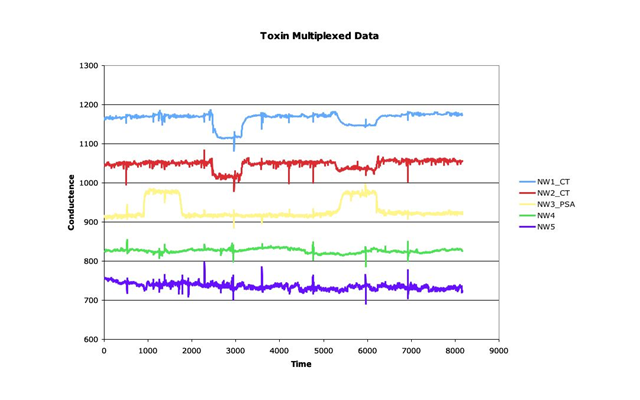}
\end{center}
\caption{Multiplexed nanowire data.}
\label{fig22}
\end{figure}

To explore the issue of correlated noise across nanowires, we consider
the data of Figure~\ref{fig10}, reproduced here for convenience as
Figure~\ref{fig22}.  Wires 1 and 2 are modified with the Cholera Toxin
(CT) antibody, wire 3 is modified with the PSA antibody, and wires 4
and 5 are modified with ethanolamine and serve as controls.

\begin{figure}[hbtp]
\begin{center}
\includegraphics[width=9cm]{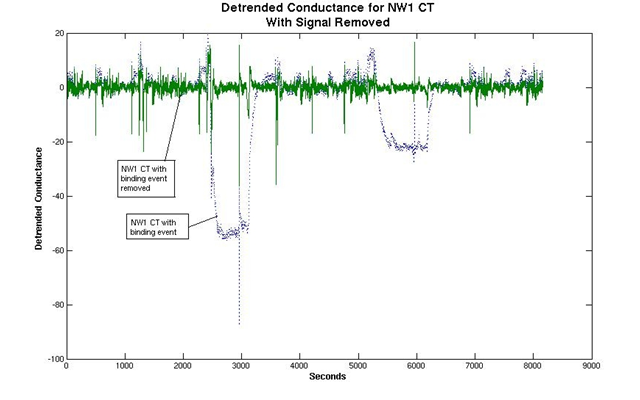}
\end{center}
\caption{Estimate of NW1 with binding signal removed.}
\label{fig23}
\end{figure}

There appear to be correlated noise events between nanowires in the
array.  If the common noise process could be extracted, this noise
process could be removed from all nanowire inputs and increase
detection performance.  To see how closely these processes are
correlated, the data in each nanowire is detrended using a moving
median filter.  Each nanowire signal is then low-pass filtered to
estimate the signal-only portion of the signal.  Subtracting the
estimated signal from the data yields an estimate of the noise-only
signal or background noise.  For example, the result of estimating and
removing the binding signature from NW1 is shown in
Figure~\ref{fig23}.  This signal-removal operation is performed for
each nanowire signal.

\begin{figure}[hbtp]
\begin{center}
\includegraphics[width=9cm]{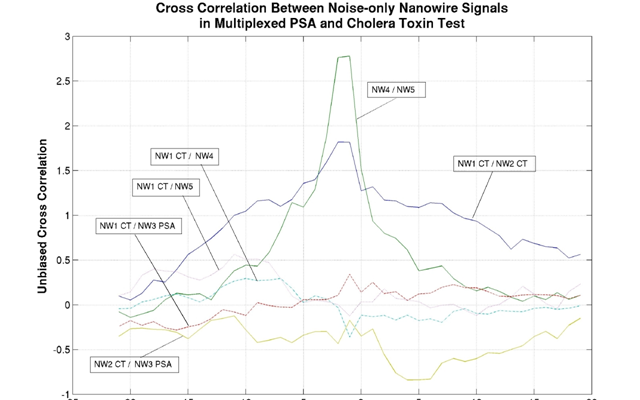}
\end{center}
\caption{Cross-correlation between noise-only nanowire signals.}
\label{fig24}
\end{figure}

Then cross-correlations between the noise-only signals for selected
nanowires are shown in Figure~\ref{fig24}.  There is a high
correlation between nanowire pairs that are similarly treated.  There
is a peak in correlation between NW1 and NW2, which were both treated
with CT antibodies.  There is also a large correlation peak between
the two control wires NW4 and NW5.

The correlations between differently treated nanowires are much lower,
but inspection shows that weak correlation peaks appear at lags that
correspond to the relative positions of spikey noise peaks.  For
instance, the lags between noise spikes in NW1 as compared to NW4 and
NW4 are about 10 seconds.  There is also a weak peak in the NW1 / NW3
cross-correlation at a lag of 1 second.  This agrees with casual spot
check of noise spike locations.

The negative-going correlation exhibited between NW2 (CT) and NW3
(PSA) indicates the different polarity of NW3 as opposed to NW2.  A
few noise spikes correspond to the peak correlation at 5 seconds lag,
but not all spikes occur at the same lag.  The correlation
distribution seems to be dominated by the low-level background noise
in this case.

This may indicate that the background noise processes have two
components -- a random noise component and one related to outside
events such as injections, changes in sample flow, electrical
interference, and so on. Similarly treated nanowires have highly
correlated backgrounds.  Differently treated nanowires have largely
uncorrelated noise backgrounds, but show some correlation that
corresponds to the lags in spikey noise.

\begin{figure}[hbtp]
\begin{center}
\includegraphics[width=9cm]{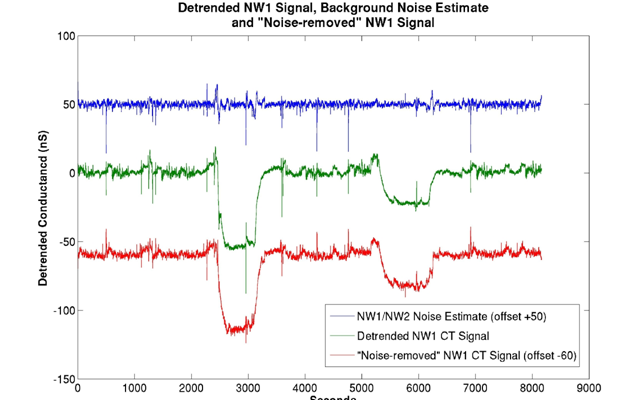}
\end{center}
\caption{Comparison of detrended NW1 signal with background noise estimate.}
\label{fig25}
\end{figure}

The correlation between similarly treated nanowires suggests that
noise background from these nanowires could be extracted, ensembled,
and subtracted from the corresponding nanowires.  In a naive
experiment, the noise-only estimates from NW1 and NW2 were averaged
with the time series offset by the peak lag (1 second -- the peak lag
was between 1 and 2 seconds).  Subtracting this simple noise estimate
from the detrended NW1 yields a ``noise-removed'' signal estimate.
The averaged noise estimate, detrended NW1 signature, and
``noise-removed'' estimate are compared in Figure~\ref{fig25}.

The variance of the detrended NW1 signal (including the binding
signature) is approximately 237.  The variance of the NW1 signal after
subtracting the noise estimate is 230.  Since the variance is
dominated by the binding signal, this shows a noticeable reduction in
noise.  In this simple example, there is some signal capture in the
noise estimate due to the filtering scheme used to remove signal.
Also, the raw noise estimate is used as opposed to determining an
optimum frequency range for noise estimation.  This approach could be
optimized with more nanowires contributing to the estimate and with
more experience with data.

\section{Future Work}

We have only encountered a limited range of response from these
arrays, and some phenomena are still not well understood (e.g., the
trend of the data in Figure~\ref{fig5}). This indicates the need for
further experimentation, exploring massively replicated arrays, the
robustness of response in such arrays, etc. especially from the point
of view of understanding the noise processes in these systems.  There
appear to be a variety of sources that we observed hints from in the
data. Designing a proper approach to the data whitening step will
require a better understanding of all the interference sources ---
this is especially true if there is a requirement for fast response as
well as a very low false alarm rate in practical applications.

\section*{Acknowledgment}

The authors thank Hung Bui for his code, and Patolsky et al. for
providing the data.

\bibliography{nano-sensing} 
\bibliographystyle{IEEEtran}
\end{document}